\pdfoutput=1

\documentclass[11pt]{article}

\usepackage{EACL2023}

\usepackage{times}
\usepackage{latexsym}
\usepackage{mathtools}
\usepackage{graphicx}
\usepackage{bbm}
\usepackage{float}
\usepackage{array,multirow}
\usepackage[ruled,vlined]{algorithm2e}
\usepackage{algpseudocode}
\usepackage{booktabs}

\usepackage[T1]{fontenc}

\usepackage[utf8]{inputenc}

\usepackage{microtype}

\usepackage{inconsolata}

\title{A User-Centered, Interactive, Human-in-the-Loop Topic Modelling System}

\author{Zheng Fang$^1$, Lama Alqazlan$^1$, Du Liu$^2$, Yulan He$^{1,3,4}$ and Rob Procter$^1$$^,$$^4$ \\
  $^1$Department of Computer Science, University of Warwick, UK\\
  $^2$Bayes Business School, City, University of London, UK\\
  $^3$Department of Informatics, King's College London, UK\\
  $^4$The Alan Turing Institute, UK \\
  \texttt{\{Z.Fang.4,~Lama.Alqazlan,~Rob.Procter\}@warwick.ac.uk}\\
  \texttt{yulan.he@kcl.ac.uk}, ~~\texttt{Du.Liu@city.ac.uk}}

\begin{document}
\maketitle
\begin{abstract}
Human-in-the-loop topic modelling incorporates users' knowledge into the modelling process, enabling them to refine the model iteratively. Recent research has demonstrated the value of user feedback, but there are still issues to consider, such as the difficulty in tracking changes, comparing different models and the lack of evaluation based on real-world examples of use. We developed a novel, interactive human-in-the-loop topic modeling system with a user-friendly interface that enables users compare and record every step they take, and a novel topic words suggestion feature to help users provide feedback that is faithful to the ground truth. Our system also supports not only what traditional topic models can do, i.e., learning the topics from the whole corpus, but also targeted topic modelling, i.e., learning topics for specific aspects of the corpus. In this article, we provide an overview of the system and present the results of a series of user studies designed to assess the value of the system in progressively more realistic applications of topic modelling.
\end{abstract}

\section{Introduction}

Huge amounts of unstructured, textual data are generated daily. As more data becomes available, it becomes more difficult to search, understand and discover the knowledge within it. Because of the human effort it requires, conventional qualitative approaches, such as Grounded Theory, \cite{glaser1968discovery} are no longer feasible with such large volumes of data. Topic modelling is a potential solution that has received increasing attention in recent research \cite{heidenreich2019media,curiskis2020evaluation,dantu2021exploratory,goyal2021measuring} to help users organize, search, and understand large amounts of information. It is an unsupervised machine learning technique for identifying hidden topics in large, unstructured text corpora, in which a hidden topic is represented by a group of words describing a common theme. Users can easily identify the topics in each document and search for documents closely associated with a specific topic for a more in-depth study. However, the topics generated by conventional topic models are often incoherent and contain many unrelated words \cite{chang2009reading,mimno2011optimizing,boyd2014care}. Although these issues can be addressed by pre-processing the target data source, for example, by removing irrelevant words from the vocabulary list and adjusting the hyper-parameters of the model, such as the number of topics, this requires familiarity with the algorithm. Hence, it is difficult for anyone who does not have some knowledge of topic modelling.  

Human-in-the-loop topic modelling (HL-TM) incorporates human knowledge into the topic modelling process to address the aforementioned issues. It allows users who are not experts in topic modelling to refine the model through a set of refinement operations, such as adding words to a topic, removing words from a topic, splitting topics, or merging topics \cite{jagarlamudi2012incorporating,wang2012semi,choo2013utopian,hoque2015convisit,lund2017tandem,smith2018closing}. While most of these studies did not feed the refinement operations into an iterative retraining process, \citet{smith2018closing} implemented a fully interactive, user-centered HL-TM system, and examined how the user experience is affected by issues arising in interactive systems, such as unpredictability, trust and lack of control.
However, there are still limitations to their work. First, their system only allows users to refine the model sequentially, meaning that once a user updates the model, a new model overrides the previous model. This prevents users from comparing the effects of applying different refinement operations to the same model, making it difficult to find the most appropriate ones. Furthermore, because the previous model is no longer accessible, users may find it challenging to decide whether the new version is really an improvement. Second, their system does not allow users to retract the changes they made after the underlying model was updated. This becomes a problem when users make inappropriate changes that lead to unexpected results. It is especially frustrating when a user has spent a lot of time refining the model and the whole effect is ruined by one inappropriate change. Third, they use Latent Dirichlet Allocation (LDA) \cite{blei2003latent} as their underlying model. 

Since LDA is a full-analysis model that can only learn topics from the whole corpus, its application is limited to when users have no \textit{apriori} assumptions about the topics of the corpus. While users can refine the model by adding prior knowledge in an attempt to turn an unrelated topic into one that focuses on the aspect of interest, the phenomenon of higher order co-occurrence in LDA \cite{heinrich2009generic} may prevent any infrequent words related to the aspect of interest being assigned to the unrelated topic. For instance, given a set of posts about health, researchers may wish specifically to analyze the impact of food on health. Researchers would add food-related words such as “food”, “eat” to an unrelated topic. If these words have a relatively low frequency of occurrence in the posts, then the system may not turn the unrelated topic into a food-related topic. Fourth, an important question that has not yet been explored is how to signal to users when their assumptions do not match reality \cite{kumar2019didn}, which may bias the refinement process. For instance, a user may think that a technology-related topic should contain words like ``apple, google", while the input corpus has the word ``apple" related to fruit. By adding ``apple" to the topic, the topic will be contaminated with fruit-related words. Such incorrect refinements would result in poor results.

To address these issues, we implemented a novel HL-TM system that supports six refinement operations, including \emph{add words, remove words, swap word order, remove documents, merge topics}, and \emph{split topic} \cite{lund2017tandem,smith2018closing}. Unlike \citet{smith2018closing}, where users can only refine the model sequentially and cannot retract the changes they made, in our system, users can make different attempts at refinement to the same model node and compare the resulting models. A complete refinement history is also presented, allowing users to track their changes from the first step. These also ensure that users can revert to the previous model node when an inappropriate refinement operation is applied. Instead of using LDA as the underlying model, our system uses the query-driven topic model (QD-TM) from \cite{fang-etal-2021-query} as the underlying model. The advantage is that the QD-TM not only supports the full-analysis capabilities from LDA, i.e., learning the overall topics from the whole corpus, but also supports users in performing targeted analysis, i.e., learning topics focused on specific aspects of the corpus by mitigating the effects of higher order co-occurrence phenomenon in LDA \cite{fang-etal-2021-query}. 

Moreover, we also implemented a novel automatic topic words suggestion feature to guide users in adding appropriate words to the selected topic. This feature extracts a list of candidate words related to the topic from which users can select words to add to the topic. Our evaluation results demonstrate the usefulness of this feature, where the suggested words are closely related to the selected topic and align better with the ground truth. We also conducted a series of user studies designed to assess the value of the system on real world application scenarios.

This work makes the following contributions: (1) a novel interactive HL-TM system with an advanced user interface that allows users to train different models from the same model node and retract inappropriate changes applied; (2) the use of QD-TM as the underlying model to support both the full-analysis and targeted-analysis topic modelling capabilities; (3) a novel and efficient topic words suggestion feature to guild users add appropriate words to the selected topics; and (4) a small scale user study and two detailed studies designed to replicate real-world application scenarios.

\section{Background}

\paragraph{Query-driven topic model} is a semi-supervised topic modelling algorithm developed by \citet{fang-etal-2021-query}. It allows users to specify a simple query in words or phrases and return query-related topics. The original model involves a two-stage modelling process; in the first stage, the model infers one topic for each concept as well as other unrelated topics, while in the second stage, the model extends the topic of each concept into a set of subtopics. In our work, we are only interested in the first stage of the model. If there is no query input, the model works as a conventional topic model.

The model is based on a variant of a Hierarchical Dirichlet Process (HDP) \cite{teh2006hierarchical}, which is a nonparametric Bayesian model that assumes that a restaurant (i.e., a document) has a set of tables and serves dishes (i.e., topics) from a global menu. A single dish is only served at a single table for all customers (i.e., words) who sit at that table. We developed our version of the algorithm based on the Gibbs sampling technique. For a word $w_{ji}$ at document $j$ and position $i$, the probability for sampling an existing table $t$ is:
\begin{equation}\small
\label{eq1}
\resizebox{\columnwidth}{!}{$
  p(t_{ji}=t \mid t^{-ji},k) \propto \mathbbm{1}(w_{ji},k_{jt}) n_{jt}^{-ji} f_{k_{jt}}^{-w_{ji}} (w_{ji})
  $}
\end{equation}
and the probability for sampling a new table $t^{new}$ is:
\begin{equation}\small
  p(t_{ji}=t^{new} \mid t^{-ji},k) \propto \alpha p(w_{ji} \mid t^{-ji},t^{new},{\bf k})
\end{equation}
Here, $t^{-ji}$ are the table assignments of all other words. $k_{jt}$ is the topic assignment of table $t$ at document $j$. $n_{jt}^{-ji}$ is the number of words in document $j$ at table $t$ and $f_{k_{jt}}^{-w_{ji}} (w_{ji})$ is the probability of $w_{ji}$ assigned to topic $k_{jt}$. $\mathbbm{1}(w_{ji},k_{jt})$ is an indicator function, which takes on value 0 if $w_{ji}$ is a pre-defined word associated to a topic $z$ that $z{\neq}k_{jt}$, and 1 otherwise. $\alpha$ is a hyperparamenter of the model. The probability for sampling a topic $k_{j{t^{new}}}$ for the new table is: 
\begin{equation}\small
  p(k_{j{t^{new}}} \mid t,k^{-j{t^{new}}})  \propto \mathbbm{1}(w_{ji},k_{jt})	{m_k} 	f_k^{-w_{ji}} (w_{ji})
\end{equation}
\paragraph{Human-in-the-Loop topic modelling} has received a lot of attention in recent years. \citet{boyer2017human} created a Human–Machine methodology for identifying Systems Thinking topics in a large corpus of text. Users are required to subjectively identify and provide seed documents describing the topic to guide the topic model's training. The methodology, however, did not incorporate any prior knowledge into the modelling process, instead it simply modified the training corpora. Various refinement operations, such as adding words, removing words, adding documents, removing documents, creating topics, merging topics, or removing topics, were implemented to better utilize human knowledge in the topic modelling process \cite{hoque2015convisit,lund2017tandem,smith2018closing}. During the model sampling process, the refinements change the prior knowledge of the underlying model. To understand the usefulness of different refinement operations, \citet{lee2017human} conducted a user-centered approach to find out how non-expert users interpret topic models and what refinement operations they want most, but they only implemented a basic system without full interaction, so the refinement operations they applied did not update the underlying topic model. 

\citet{smith2018closing} took one step forward from Lee’s work by implementing a fully interactive, user-centered HL-TM system and further examined how common interactive machine learning challenges, such as unpredictability, latency and trust, affect the user experience. \citet{kumar2019didn} were the first to comparatively evaluate different refinement operations, as well as two feedback injection frameworks, namely informed priors and constraints. They not only suggested that future research should test the system with end users, since their experiments only used simulated user behavior, but also mentioned that it’s important to signal to users when their assumptions do not match reality. Though other HL-TM systems exist such as UTOPIAN \cite{choo2013utopian} and ConVisIT \cite{hoque2015convisit}, they use alternative approaches such as non-negative matrix factorization and fragment quotation, and do not provide the complete set of refinement operations that users need \cite{lee2017human}.

Our work can be seen as an extension of the work of \citet{smith2018closing}. Compared to these studies, it not only provides a fully interactive HL-TM system incorporating various refinement operations but also provides a more user-centered design, such as a complete refinement history and a topic words suggestion feature to enhance the user experience.

\section{Proposed System}

We introduce our implementation and the interface design in this section. 

\subsection{Refinement Implementations}
Our system uses Gibbs sampling as the inference technique and adopts the constraint method described in \citet{kumar2019didn} to inject new information. Every time a user provides a feedback to the system, it first forgets table-word assignment $t_{ji}$ and then injects new information into the system using a potential function $f(k,w,j)$ \cite{yang2015efficient} of the hidden topic $k$ of word $w$ in document $j$. The equation (\ref{eq1}) then becomes:
\begin{equation}\small
\resizebox{\columnwidth}{!}{$
  p(t_{ji}=t \mid t^{-ji},k) \propto \mathbbm{1}(w_{ji},k_{jt}) n_{jt}^{-ji} f_{k_{jt}}^{-w_{ji}} (w_{ji})f(k_{jt},w_{ji},j)
  $}
\end{equation}

Prior work \cite{lee2017human,smith2018closing}  discovered that users typically prefer simple refinement operations, therefore, we implemented the following six refinement operations that are commonly used by users in previous studies: 

\paragraph{Add word} $x$ to topic $z$. We update the potential function $f(k,w,j)$ such that $f(k,w,j) = 1$ if $k=z$ and $w=x$, otherwise it is assigned a value of 0.
\paragraph{Remove word} $x$ from topic $z$. We update the potential function $f(k,w,j)$ such that $f(k,w,j) = 0$ if $k=z$ and $w=x$, otherwise it is assigned a value of 1.
\paragraph{Swap word order} of $w_1$ and $w_2$ in topic $z$ so that $w_2$ has higher order than $w_1$. We first compute the ratio $r$ between the difference $n_{w_1,z}-n_{w_2,z}$ and $n_{w_2}$, where $n_{w_1,z}$ and $n_{w_2,z}$ are the counts of $w_1$ and $w_2$ in topic $z$, respectively, and $n_{w_2}$ is the counts of $w_2$ in all topics except $z$. We then update the potential function $f(k, w, j)$, such that $f(k, w, j) = 1$ if $k = z$ and $w = w_2$, otherwise it is assigned $\delta$, where $\delta = 0$ if $r > 1$, otherwise $\delta=1.0-r$. 
\paragraph{Remove document} $d$ from topic $z$. We update the potential function $f(k,w,j)$ such that $f(k,w,j) = 0$ if $k=z$ and $j=d$, otherwise it is assigned a value of 1.
\paragraph{Merge topic} $t_2$ into $t_1$. In the next Gibbs sampling iteration, we only sample topics for words assigned to $t_2$ in the previous Gibbs sampling iteration. We update the potential function $f(k,w,j)$ such that $f(k,w,j)=1$ if $k=t_1$ and $w \in t_2$, otherwise it is assigned a value of 0.
\paragraph{Split topic} $t$ into two topics using seed words $s$, i.e., $s$ need to be moved from $t$ to a new topic $t_n$. To do this, we first create a new topic using the nonparametric model. Then, we apply the \emph{add word} operation to add all the words in $s$ to $t_n$.

\subsection{Topic words suggestion}
We developed a novel topic words suggestion feature in QD-TM to let users decide which words should appear or not appear in the topic words. Using only subjective input may result in topic words that are spurious, but combining the suggested words can help  users to steer the model in the right direction. The topic words suggestion feature is integrated into the Gibbs samplings process of the model. It has two stages. First, it samples an indicator of whether a document is relevant to a topic or not. Second, it updates the suggested words from the relevant documents. 

\paragraph{Document relevance sampling} For each topic $k$, if a document contains any of the suggested words and is sampled to be relevant to the topic in the previous Gibbs sampling iteration, then it is assumed to be relevant to the topic in the current Gibbs sampling iteration. Otherwise, we use the following equation modified from \citet{wang2016targeted} to decide if the document is relevant to the topic: 
\begin{align}\small
p(r_k=c\mid \bf{r},\pi,\beta,\gamma_r,\gamma_{ir})\propto\nonumber\qquad\qquad\qquad\qquad\qquad\qquad\\
\begin{cases} 
(C_c^{R(-m)}+\gamma_r)\times\frac{\prod_{v}^V\Gamma(n_{k,v}^{-m}+\beta)}{\Gamma(\sum_v^V(n_{k,v}^{-m}+f_{c,m,v})+V\beta)}  \\ 
\mbox{if }c=1 \\
(C_c^{R(-m)}+\gamma_{ir})\times\frac{\prod_{v}^V\Gamma(n_{\tilde{k},v}^{-m}+\beta)}{\Gamma(\sum_v^V(n_{\tilde{k},v}^{-m}+f_{c,m,v})+V\beta)} \\
\mbox{if }c=0 
\end{cases}
\label{eq5}
\end{align}

\noindent where $C_c^{R(-m)}$ is the number of documents under relevance status $c$ excluding the current document $m$, $n_{k,v}^{(-m)}$ is the counts of term $v$ in target topic $k$ excluding the words from the current document $m$, $f_{c,m,v}$ is the frequency of term $v$ in document $m$ under relevance status $c$, and $\pi$ indicates the Bernoulli distribution over relevance status. $\beta$, $\gamma_r$, and $\gamma_{ir}$ are hyperparameters, and $V$ is the vocabulary size of the dataset. 

\paragraph{Automatic keywords expansion} To get suggested words for a topic in each Gibbs sampling iteration, we first calculate the score of term $v$ in the relevant documents of the topic:
\begin{equation}\small
 Score(v)=P_R(v)log\frac{P_R(v)}{P_C(v)},
\end{equation}
where $P_R(v)$ is the probability of term $v$ in the relevant documents and $P_C(v)$  is the probability of term $v$ in the whole corpus. We extract terms with high scores as the candidate terms. We then calculate the cosine similarity between each candidate term and the embedding of the topic, and only add terms to the suggested words list when the similarity is greater than 0.5. The embedding of a topic $k$ can be obtained by:
\begin{equation}\small
 emb(k)=\sum_{m}^{M}P^i(k,m)emb(m),
\end{equation}			
where $p^i(k,m)$ is the probability of $m^{th}$ word of topic $k$ in $i^{th}$ iteration, $M$ is the total number of representative words in the topic, which is usually set to 10, and $emb(m)$ is the pretrained word embedding of the $m^{th}$ word.

\subsection{Interface Design}

The user interface of the system consists of two windows (Appendix A). In the first window (Figure \ref{Fig.main1}), users can define the hyper-parameters of a topic model, such as the initial number of topics. If users are interested in topics describing specific concepts or aspects of the corpus, they can also define the prior knowledge of a topic model in this window. If no prior knowledge is provided, then the model behaves as a conventional topic model without any prior knowledge.

The second window (Figure \ref{Fig.main2}) displays the detailed information of the trained model. Users can view the model as a list of topics. Users can also refine the model using the refinement operations implemented and track model change histories. Different from previous systems where users can only refine a model sequentially and cannot retract the changes they made, we include a model history panel with a novel model tree structure so users can make different attempts at refinement to the same model node and compare the resulting models. To help users interpret each topic, we also show a set of topic labels from the automatic topic labeling algorithm \cite{mei2007automatic}, where a topic label is a phrase that summarises the main idea of the topic. A detailed description of the interface can be found in Appendix A.

\section{Evaluation}


The evaluation is divided into two parts. In the first part, we evaluated the performance of the topic words suggestion feature in a controlled, laboratory experiment, where we compared two versions of our system. In the second part, we evaluated the HT-LM system by applying it in two realistic topic modelling use cases. 

\subsection{Topic words suggestion evaluation}
\paragraph{Datasets} Three commonly used datasets for topic modelling were chosen: the 20newsgroup \footnote{http://qwone.com/~jason/20Newsgroups/} dataset containing 18k news posts from 20 categories; the TagMyNews \footnote{http://acube.di.unipi.it/tmn-dataset/} dataset containing 32k short English news from 7 categories; and the SearchSnippets \cite{xu2017self} dataset containing 12k short English news from 8 categories.

\paragraph{Baselines} Our HL-TM system uses QD-TM as the underlying model. To test whether the \emph{topic words suggestion} feature can improve the original model, we used only this feature to refine the QD-TM and compared the refined model with the original QD-TM.

\paragraph{Parameterisation} We focused on the targeted topic modelling capabilities of QD-TM. To make fair comparisons, we adopted the same experimental settings described in \citet{fang-etal-2021-query}. We used query phrases to represent the main concept of each category in a dataset, following the same setup as in 
\citet{fang-etal-2021-query}. Categories that do not have meaningful names were removed from the datasets, e.g., \texttt{talk.politics.misc} in the 20Newsgroup dataset. The query phrases were integrated into the model using the first window of our HL-TM interface.

We used our HL-TM system to refine the original QD-TM and only used the \emph{add word} refinement operation. In each Gibbs sampling iteration, for each topic, we added all suggested words to the topic. 
For the hyper-parameters in equation (\ref{eq5}), $\beta$ is set to 0.5, and both $\gamma_r$ and $\gamma_{ir}$ are set to 1. We ran 2000 Gibbs sampling iterations.

\begin{table*}[htb]
\centering
\resizebox{0.96\textwidth}{!}{
\begin{tabular}{lccccccccc}
\toprule
\bf Model & \multicolumn{2}{c}{20news} & \multicolumn{2}{c}{TagMyNews} & \multicolumn{2}{c}{SearchSnippets}  \\ 
\cmidrule(lr){2-4} \cmidrule(lr){5-7} \cmidrule(lr){8-10}
 & Coherence & Precision@K & Coherence & Precision@K & Coherence & Precision@K\\  \midrule
{\bf QD-TM } & 0.445 & 0.612 & 0.413 & 0.736 & 0.475 & 0.741  \\
{\bf QD-TM + words suggestion  } & {\bf 0.482} & {\bf 0.634} & {\bf 0.477} & {\bf 0.763} & {\bf 0.481 } & {\bf 0.825 }  \\
\bottomrule
\end{tabular}}
\caption{\label{font-table} Average results for adding suggested topic words for model refinement. We ran each model five times. We used the topic coherence metric C\_V from \citet{roder2015exploring}}
\label{Table.main1}
\end{table*}

\paragraph{Results} We evaluated the quality of the final models in terms of topic coherence and document retrieval performance as in \citet{fang-etal-2021-query}. Better precision@K scores indicate that the learnt topics are more discriminative and representative.
Higher coherence scores indicate better topic interpretability. Table \ref{Table.main1} shows the performance of adding suggested topic words to the model. 
We observed that the refined model achieves higher coherence scores compared with the original model. It indicates that the suggested words are semantically related to each topic and can help produce more coherent topics. The better precision@K scores also show that the suggested topic words are highly relevant to the predefined categories of the dataset. 
This is expected as the word suggestion feature in our system tends to identify more important words related to the category of the corresponding topic. 
Instead of blindly adding the suggested words to the model, our system allows users to decide which suggested words should be added. This allows them to use their domain knowledge to further filter out noise. 

\subsection{Laboratory evaluation}
\paragraph{Description} To compare our system with the state-of-the-art human-in-the-loop topic modelling system \cite{smith2018closing}, we recruited 20 participants via a university campus email list to run a small-scale user study, with one subject group working with an equivalent of \citet{smith2018closing} -- our system with no model history, no words suggestion, and no topic labeling (old system) -- and one group working with our full-featured system (new system). In a factorial design with two independent variables, we used two corpora of Reddit posts focusing on online teaching platforms (corpus A) and Twitter tweets discussing the 2021 United Nations Climate Change Conference (corpus B). Participants were randomly allocated into two groups. The first group did corpus A with the old system and corpus B with the new system, and the second group did corpus B with the old system and corpus A with the new system. The task was to conduct a qualitative analysis of the datasets.

Participants randomly started with either the old system or the new system to eliminate the influence of training effects. We then evaluated the average topic quality of the two system conditions. In addition to comparing topic quality across systems, we also asked participants questions about how much they like the new features. All participants were fluent English speakers and non-experts in topic modelling. Each participant received a £20 Amazon gift card as payment for the experiment.

\paragraph{Dataset and Topic Model} We used two datasets for the experiments. The Reddit dataset contains 9,651 posts focusing on online teaching platforms. These posts were randomly sampled from the original dataset taken from \citet{alqazlan_procter_castelle_2021}. The Twitter dataset contains 8,990 tweets related to the 2021 United Nations Climate Change Conference. We used keywords related to the conference to search for tweets between 31 October 2021 and 17 November 2021. These keywords (see Table \ref{Table.main2}) were checked and provided by social scientists researching COP26 climate change tweets through a few rounds of manual inspection of the tweets. We used QD-TM as the underlying model, and used a standard stop words list and 2000 Gibbs sampling iterations to initialize the topics. 
We set the initial topic number to 10 and 13 respectively for the Reddit and Twitter datasets, respectively, based on our previous experience with these datasets. For each subsequent update during the task, 10 Gibbs sampling iterations were run.

\paragraph{Procedure} Sessions took around 60 minutes on average, and they were conducted face-to-face. We began by introducing participants to topic modelling and how to use the tools to refine topics. For both systems, we asked them to first read the top five posts of each topic to interpret the underlying theme and then read the corresponding top 10 topic words and use the provided refinement operations to refine the topic until the top 10 topic words and the top 5 posts are consistent with the interpreted theme. For the new system, participants could also access the new features -- model history, word suggestions, and topic labelling to assist their in refinements. We asked participants to click the \textit{apply refinements} button to update the underlying model after they have added all the necessary refinements to a topic. They were allowed to undo any operations that have not been applied to the underlying model. After they finished the refining tasks, we asked them to rate how much they are satisfied with the resulting topics and how much they like the new features. Due to time limitations, we only asked participants to refine the first five topics of each system. The first five topics of the starting model for each 
corpus can be found in Table \ref{Table.main3} and Table \ref{Table.main4}.

\paragraph{Findings} We recorded user interactions with the tools. Users performed a total of 1,284 refinements using the two systems. Among the top four most used operations were the \textit{delete words} operation (used 554 times), the \textit{swap words} operation (used 512 times), the \textit{add words} operation (used 171 times), and the \textit{delete document} operation (used 81 times). This is consistent with the findings of \citet{smith2018closing}, who also observed that these operations were the four most used operations (excluding the add to stop words operation since our system didn't implement this). For the new system, the \textit{add words} operation was applied 115 times in total, among which 56 times were adding suggested words, which indicates that the suggested words operation had high usage. We report the usage of refinements for each subject group in Table \ref{Table.main5}.
    
To evaluate whether using the new system can result in better topics than the old system, participants were asked to rate their satisfaction with the final topics of the two systems on a five point scale (1 not very satisfied, 5 very satisfied). For the Reddit dataset, the average satisfaction score for the new system was 4.2 (SD=0.67), while the score for the old system was 3.89 (SD=0.78).  Although a Mann-Whitney U test (U=30.5, z=0.83887, p=.200) showed this difference was not statistically significant, 8 out of 10 participants were satisfied with final topics of the new system. For the Twitter dataset, the score for the new system was 4.0 (SD=0.71), while the score for the old system was 3.2 (SD=0.67). A Mann-Whitney U test (U=18.5, z=1.8985, p=.029) showed that the difference was statistically significant, suggesting that the new system can help improve topic quality. 

To evaluate whether the model tree feature can help users track changes, we asked participants to rate their agreement with the statement, ``I was able to remember what the model looked like before my updates'', on a five point scale (1 strongly disagree satisfied, 5 strongly agree), for both the old and new systems. The average agreement was 3.1 (SD=1.07) for the old system and 3.9 (SD=0.85) for the new system. A Mann-Whitney U test (U=116, z=2.25868, p=.012) showed the difference to be statistically significant. Participants were also asked to rate their agreement with the statement, ``the model tree feature of the new system can help me track my updates''. The average agreements for the statement was 4.35 (SD=1.04). This strongly suggests that the model tree feature of the new system can help users track their changes. To evaluate the usefulness of the words suggestion feature, we asked participants to rate their agreement with the statement, ``the suggested words feature of the new system can help me identify relevant words of the topics''. The average agreements for the statement was 4.1 (SD=0.64). 17 out of 20 participants stated that the feature is very useful, suggesting that the feature is a good supplement for adding words to topics. The finding is consistent with the observation of user interactions with the new system that around 50\% (56 out of 115) of added words were from the suggested words list. For the topic labelling feature, participants were asked if they agreed with the statement, ``the topic labels from the full-featured system can help me interpret the meaning of the topics''. The average score for the statement was 3.25, and 11 out of 20 mentioned that the provided labels for some topics are totally irrelevant. This suggests that the algorithm \cite{mei2007automatic} doesn't fit the datasets well and suggests the needs for a more accurate topic labelling algorithm in the future. 

\paragraph{Suggestions} Though the user studies show promising results, participants also had suggestions for improving the new system. Six mentioned that the swap words operation didn't fit their needs well. Instead of swapping the order of the two selected words, they prefer to allow the selected words to be inserted in new positions, which provides more flexibility when ranking topic words. One participant suggested allowing the undo any of the previously added operations directly, rather than starting with the most recently operation. 

\subsection{Use case one: Tutors’ experiences in the Gig Economy}
\paragraph{Description} To evaluate our system on real-world document analysis tasks, we invited a researcher who is investigating the Gig Economy to use our model. In previous work, \citet{alqazlan_procter_castelle_2021} applied LDA to identify topics that are related to tutors’ experiences in the Gig Economy from a Reddit dataset where tutors posted about their experiences of working on online teaching platforms. 
We asked the researcher to use our system to identify the relevant topics in the dataset and assess whether using our system can produce better results than LDA. 
As \citet{alqazlan_procter_castelle_2021} has previously found that a 17-topic LDA extracted the most number of relevant topics, for a fair comparison, the initial number of topics for our system was also set to 17. The researcher was encouraged to use any refinement operations provided by the system until she felt that the top 5 relevant posts of each detected topic revolved around one main theme.

\paragraph{Results} A total of 16 models were trained and two branches were created during the refinement process, with models 11, 12 and 13 constituting one branch and models 14, 15 and 16 constituting the other branch. Both of these branches were extended from model 10, which was trained sequentially from the initial model. By using the model tree panel to compare these models, it was found that Model 16 produced the most satisfactory results. Of all the refinement operations, \emph{remove document} was used the most and \emph{swap word order} was used the least. The researcher commented that the \emph{swap word order} could be more useful if it only adjusts the position of a word in the topic, rather than swapping a word with another one. 

We compare LDA with Model 16 and present the qualitative results here. For both models, the researcher was asked to manually assign a label to each topic. 
For the LDA model, 11 topics were found to be relevant to tutors' experience in the Gig Economy, two of which (Table \ref{Table.main6}) were missed by Model 16, while for Model 16, 16 topics were found to be relevant, of which seven topics (Table \ref{Table.main7}) were not identified by the LDA. 
This indicates that our HL-TM system is able to assist the researcher to identify more relevant topics.

To determine whether using our system can help produce better quality topics, we measured the topic coherence score for the 9 relevant topics identified by both LDA and Model 16 (Table \ref{Table.topics}). We used the best performing topic coherence measure CV based on the external corpus (Wikipedia) \cite{roder2015exploring}. We found that our system produced better topic coherence scores for six topics, while the LDA produced better scores for only three topics.
This finding is consistent with the researcher's view that only two topics from the LDA were of better quality than Model 16. We present the relevant topics from both LDA and Model 16 in Appendix C.

\subsection{Use case two: Patenting strategies for pre-determined patent value categories}

\paragraph{Description} We invited a second researcher, who is working on patenting strategies, to evaluate our system. In a previous study, \citet{ribeiro2020private} has identified a set of patent value categorises (Table \ref{Table.main8}) for synthetic biology patents. However, their work was based entirely on human evaluation and was, therefore, time-consuming, and biased evaluations could occur, so only 102 patent documents were analysed.
In this use case, the researcher aimed to use our system to categorise a larger patent sample to the pre-defined categories and reduce human selection bias.
2607 related patents in the United States Patent and Trademark Office (USPTO) dataset were used. 
The pre-defined categorises are presented in Appendix D.

To identify documents in their predefined categories in the dataset, the researcher used the first window of the interface to incorporate prior knowledge. A list of final concept words of each category is presented in Table \ref{Table.main9} in Appendix D. The initial number of topics is set to 20. 

\paragraph{Results} By using the first window of the user interface to incorporate priori knowledge, the initial model was able to roughly infer topics for the pre-defined categories, although the topics were not yet of high quality. 
We then asked the researcher to examine the top 10 relevant documents of these topics to assess the categorisation quality. The average precision is 0.3. The researcher then refined these topics using the system until she felt that it best categorised the top 5 relevant documents into each category and was satisfied with the resulting topic words. In total 23 models were trained and five branches were created, with Model 23 being the best one. After the refinements, we asked the researcher to further examine the top 10 relevant documents of the refined topics to compare with the initial model. The average precision for the refined topics is 0.96, which is much higher than the result from the initial model. This verifies that the use of the system can help identify a larger sample of documents in pre-defined categories than the previous manual evaluation method. In total, the \emph{add word} operation was applied 62 times, among which 41 times were related to adding suggested words from the system. This shows that our topic words suggestion feature can indeed identify words that are highly relevant to specific topics.

We also compared the topic coherence scores for the focused topics from the initial model (Table \ref{Table.main9}) and from Model 23 (Table \ref{Table.main10}). The scores are 0.404 and 0.434, respectively. It shows that using the system can help produce better quality topics.

\section{Conclusions and Future work}

We have developed a novel user-centered, interactive, HL-TM system to address the limitations of the prior work \cite{smith2018closing}. An advanced user interface with a model history panel is presented to allow users compare different models and retract inappropriate changes. The use of the QD-TM as the underlying model supports both the full-analysis and targeted-analysis capabilities. A novel topic words suggestion feature is also integrated into the system and the evaluation shows that the feature is promising for suggesting reasonable and coherent words for topics. A small scale experimental user study and two detailed use cases further verified the usefulness of the system on real-world tasks. From the use cases, we observed that both researchers trained many different models with multiple branches created (use case one has 16 models trained and two branches created, while use case two has 23 models trained and five branches created). This shows that researchers are very likely to change their mind when refining a topic model, so allowing researchers to compare different models and retract their changes is helpful. In the future, we plan to improve the system based on the results from the user evaluations and conduct more extensive evaluations to gather feedback on deploying the system in a wide variety of applications.


\section*{Limitations}
We used two small datasets in the user evaluations, so participants were not affected by latency issues, where the user would have to wait while the algorithm performs updates. Since QD-TM is more complex than LDA, and the words suggestion feature further increased the complexity of the computation, a large dataset could lead to latency problems. The time complexity of our system is O(DLK + KV), where D is the number of documents used, L is the average document length, K is the number of topics and V is the vocabulary size. We tested the system with two different dataset sizes (20,000 tweets vs. 8,990 tweets). By using 20,000 tweets, the average wait time for model updates during user interaction (10 Gibbs sampling iterations) was 34 seconds, compared to 14 seconds using 8,990 tweets. 

According to \citet{smith2018closing}, longer wait times can negatively affect users when using interactive systems, suggesting that latency could be an issue in our system as the size of the dataset increases. However, as is well known, these effects can be mitigated through the provision of time affordances \cite{conn1995time}. We will address this in the future.

In addition to the latency issue, the laboratory evaluation we conducted also has limitations. Not all the participants were familiar with qualitative analysis. It is likely that participants with extensive experience in qualitative analysis identify more valuable refinements to the models, resulting in better quality topics. 

\section*{Acknowledgements}
This work was supported in part by the UK Engineering and Physical Sciences Research Council (grant no. EP/V048597/1). YH is supported by a Turing AI Fellowship funded by the UK Research and Innovation (EP/V020579/1). ZF receives the PhD studentship jointly funded by the University of Warwick and China Scholarship Council.

\bibliography{anthology,custom}
\bibliographystyle{acl_natbib}

\clearpage
\appendix

\section*{Appendix A: Interface Design}
\label{sec:appendix}

\setcounter{figure}{0}
\renewcommand{\thefigure}{A\arabic{figure}}

\begin{figure*}[p]
\centering
\resizebox{0.85\textwidth}{!}{
\includegraphics[width=1\textwidth]{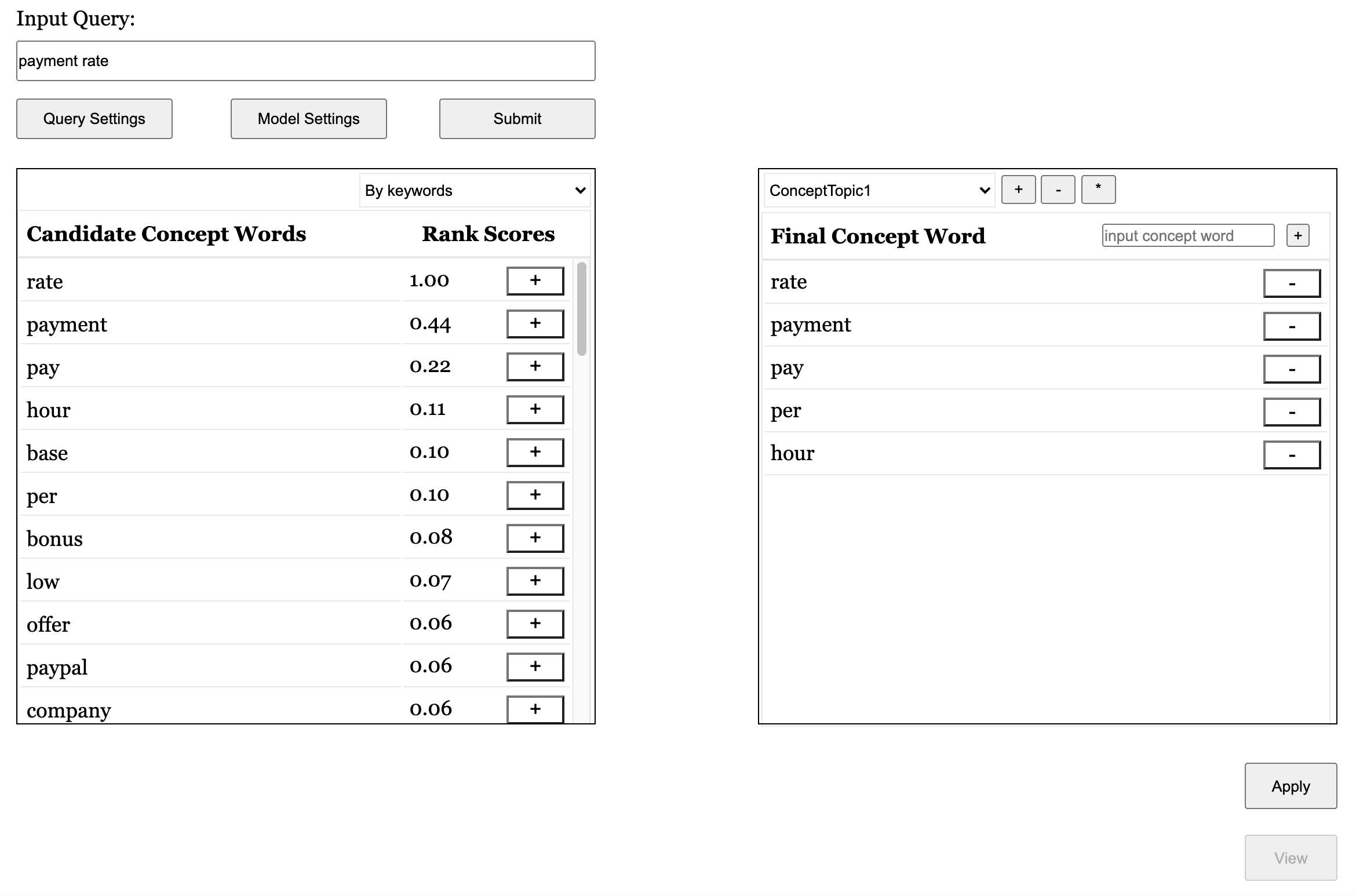}
}
\caption{The first window of the user interface. Users can define the prior knowledge of a topic model here.}
\label{Fig.main1}
\end{figure*}

\begin{figure*}[htb]
\centering
\resizebox{0.85\textwidth}{!}{
\includegraphics[width=1\textwidth]{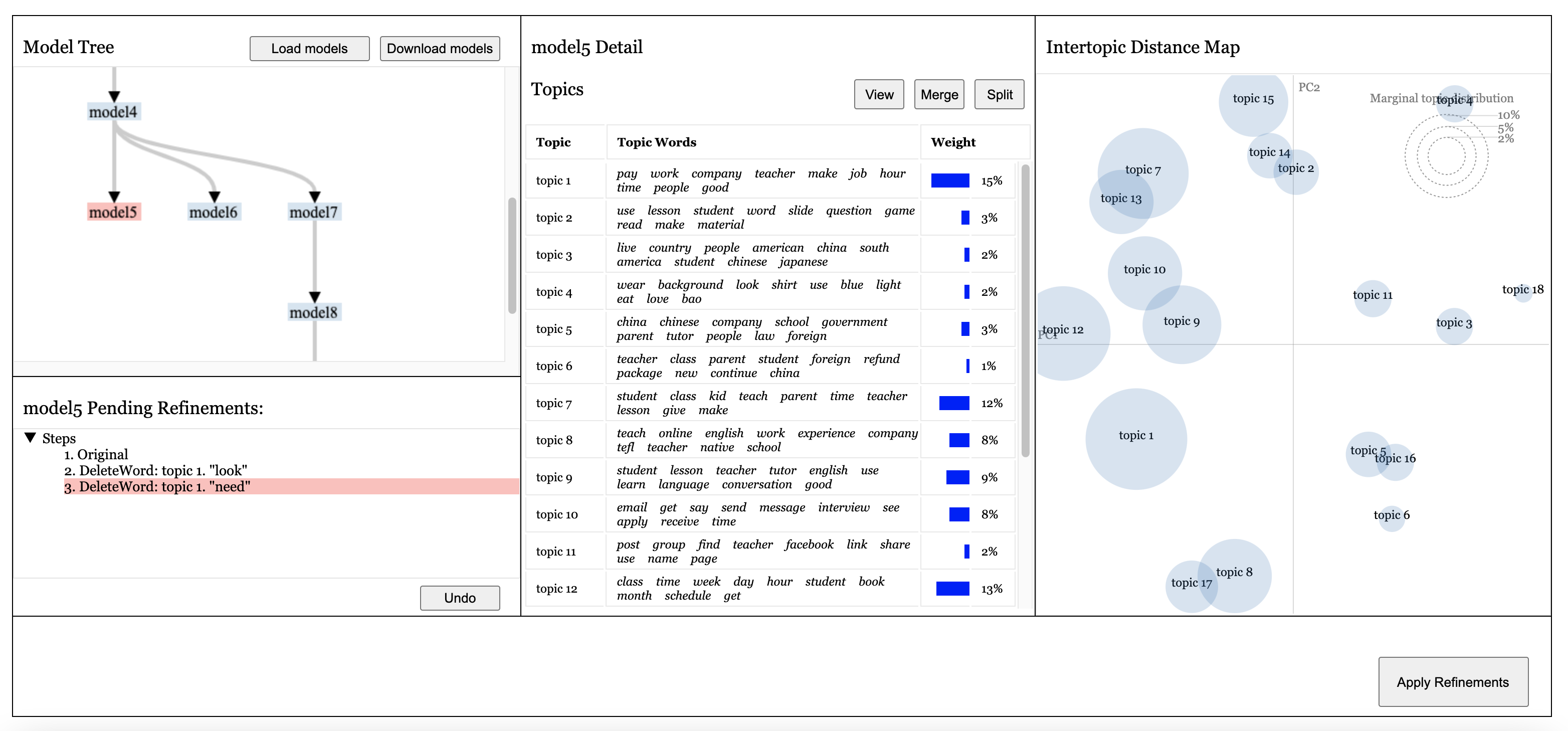}
}
\caption{The second window of the user interface, where users can view the details of the selected model and apply merge or split topics refinement operations. Users can also compare different models that have been previously refined and keep track of their previous changes.}
\label{Fig.main2}
\end{figure*}

The user interface of the system consists of two windows. If users are interested in topics describing specific concepts or aspects of the corpus, they can define the prior knowledge of a topic model in the first window (Figure \ref{Fig.main1}). They can use a query phrase to define the concept of interest. The input query phrase is expanded to a set of candidate concept words using the concept words extractor from \citet{fang-etal-2021-query}. Users can use their knowledge to determine which words should be included in a topic and click the "+" buttons next to them to add to the final concept words list of a topic. They can also add words that are not in the candidate words list to the final list based on their knowledge. The words in the final list can be removed by clicking the corresponding “-” buttons. It also supports viewing the top relevant documents of the input query by changing the viewing options from "By keywords" to "By documents".

If users have no preferred interest in the topics of the corpus, they can leave the final concept words list of each topic empty, and the model behaves as a conventional topic model without any prior knowledge. Users can click the "apply" button to apply the settings to the underlying model. Users can view the resulting topics by clicking the "View" button in the bottom, which takes them to the second window of the user interface.

The second window of the interface (Figure \ref{Fig.main2}) has three panels: \emph{model history} panel (left panel), \emph{model detail} panel (middle panel), and \emph{inter-topic distance map} (right panel). The \emph{model history} panel consists of two subpanels: the \emph{model tree} panel (top panel), which displays the refined models in a tree structure, and the \emph{refinements history} panel (bottom panel). 
Every time a user refines a selected model, a new model is added to the tree. As shown in Figure \ref{Fig.main2}, users can also create new branches from the same model node. For example, model 5, model 6 and model 7 are all refined from model 4. The \emph{refinements history} panel allows users to view the refinements history between two connected models by clicking on the edge between them. Users can also view the pending refinements of the selected model as shown in Figure \ref{Fig.main2}. 

The pending refinements will not be applied to the underlying model until users click the "Apply Refinement" button. We also include an "undo" button to allow users to undo previously added refinements. The \emph{model history} panel allows users to compare previously refined models and keep track of their previous changes to further assist the refinement process. By clicking the "Download models" button, users can download the entire model tree to their local machine, and by clicking the "Load models" button, users can load a previously downloaded model tree to continue the refinements.
 
\begin{figure*}[htb]
\centering
\resizebox{0.85\textwidth}{!}{
\includegraphics[width=1\textwidth]{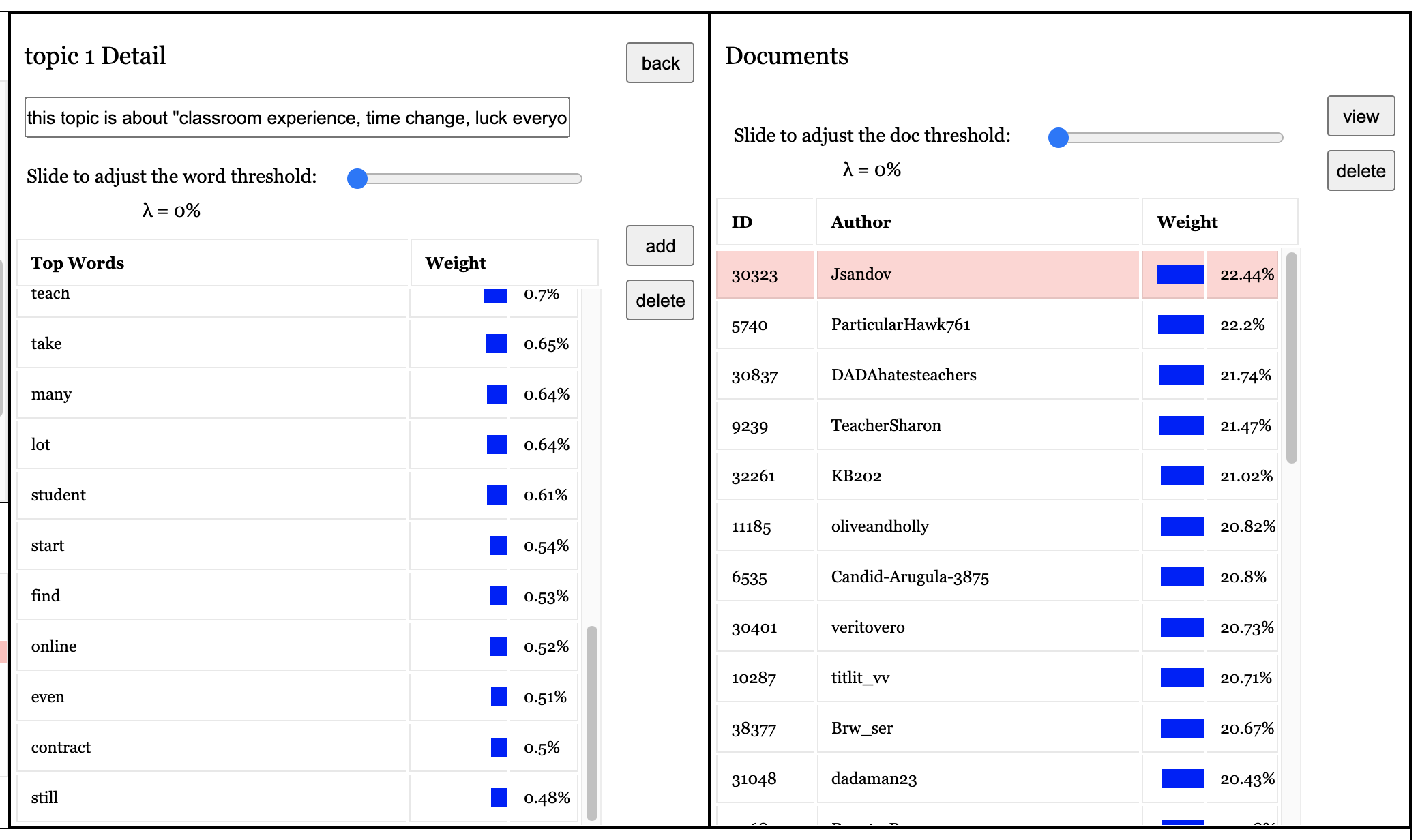}
}
\caption{The specifics of a selected topic. Users can apply add, remove or swap word order refinement operations here.}
\label{Fig.main3}
\end{figure*}

The \emph{model detail} panel, shown in the middle of Figure \ref{Fig.main2}, provides an overview of the selected model's topics, as well as the top words in each one. Weight represents the prevalence of the topic in the whole corpus. Users can rename a topic by typing a new name in the “Topic” column. Users can also merge or split topics in this panel. The right side of Figure \ref{Fig.main2} is the \emph{inter-topic distance map}. Each circle represents a topic, with its size representing the topic weight in the corpus. The \emph{inter-topic distance map} is an intuitive way to reveal the quality of a topic model where a larger distance between topics indicates a better model \cite{sievert2014ldavis}. By clicking the “view” button on the window, 
the selected topic's specifics will then be presented as shown in Figure \ref{Fig.main3}.

The top left side of Figure \ref{Fig.main3} shows a set of topic labels from the automatic topic labeling algorithm \cite{mei2007automatic}, where a topic label is a phrase that summarises the main idea of the topic. The aim of the topic label is to help users interpret the topic. We also present the top words of the topic with corresponding topic-word weights. Weights indicate the prevalence of the words in the topic. 
Users can apply \emph{add, delete} or \emph{swap word order} refinement operations here. The right side of Figure \ref{Fig.main3} displays the top documents associated with the topic and ranks them based on their weight, representing the document's prevalence of the topic. 
Users can apply the \emph{remove document} refinement operation here. By clicking the "view" button, users can view the details of the selected document.

\begin{figure*}[htb]
\centering
\resizebox{0.65\textwidth}{!}{
\includegraphics[width=1\textwidth]{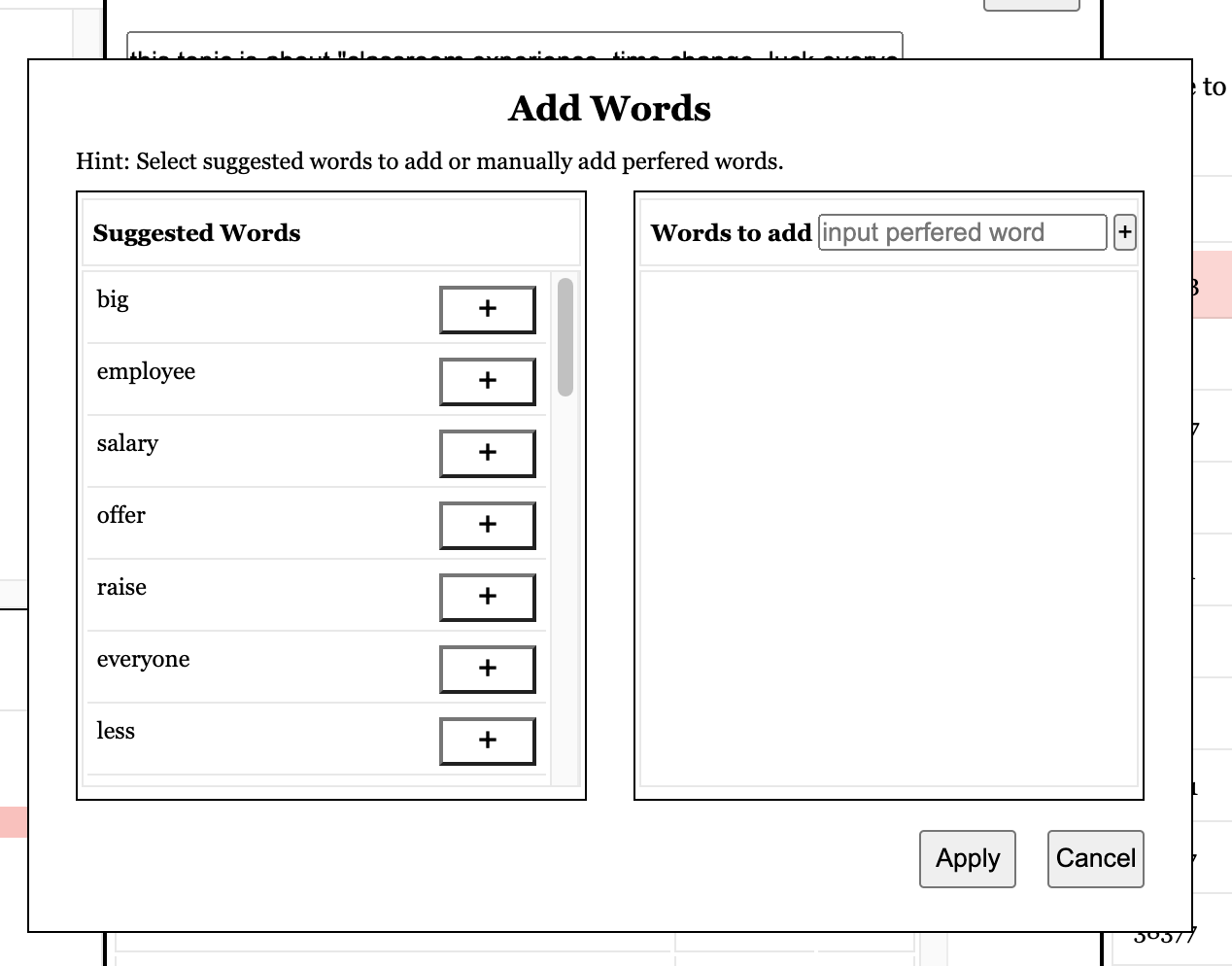}
}
\caption{The “add words” sub-panel.}
\label{Fig.main4}
\end{figure*}

When users click the "add" button in Figure \ref{Fig.main3}, an "add words" sub-panel (Figure \ref{Fig.main4}) appears. A list of words suggested by the topic words suggestion feature is displayed in the panel. Users can select from the suggested words or from their own knowledge to add words to the selected topic.

\clearpage
\onecolumn
\section*{Appendix B: Tables for the small-scale user study (section 4.2)}
\label{sec:appendix2}

\setcounter{table}{0}
\renewcommand{\thetable}{B\arabic{table}}

\begin{table}[htb]
\centering
\resizebox{0.8\textwidth}{!}{
\begin{tabular}{p{12cm}}
\toprule
{\bf keywords} \\
\midrule
police liaison, Police Liaison Officer, PLO, blue bib, \#peacefulprotest, \#righttoprotest, Police, Policing, Anarchy, Anarchism, Violence, \#ClimateJustice, \#ClimateCrisis, \#ClimateAction, cop26, \#GlobalDayofAction, \#GlobalDayforClimateJustice, \#greenwash, Pigs, Fascist, Stormtrooper, Nazi, crowd, march, rally, mob, extremist, class, privilege, eco-zealot, vandal, nutter, lunatic, eco-fascist, hypocrite, far-left, chaos, stunt, marxist, terror, rabble, anarchist, filth, rozzer, clown, pc plod, old bill, polis, wankers, woke, eco-warrior, campaign, radical, extremist, zealot, authoritarian, tyranny, tyrant, numpty, numpties, scum, disgusting, awful, unbelievable, evil, frightening, selfish, vicious, violent, thug, brutal, vicious, fight, carnage, blood, injury, hostility, aggression, force, assault, invasion, offensive, friction, stress, strain, damage, hurt, harm, block, kettle, contain, arrest, imprison, charge, thankyou, thanks, carnival, festival, fun, enjoy, party, tension, tense, disrupt, soft  \\
\bottomrule
\end{tabular}}
\caption{\label{font-table} Keywords used to retrieve tweets related to the 2021 United Nations Climate Change Conference.}
\label{Table.main2}
\end{table}

\begin{table}[htb]
\centering
\resizebox{0.8\textwidth}{!}{
\begin{tabular}{p{2cm}p{12cm}}
\toprule
{\bf Topic} & {\bf Top 10 words} \\
\midrule
topic 1 & hour class week pay month work time day start make \\
topic 2 & email apply work company good check send interview hire process \\
topic 3 & teach english online native tefl degree experience company teacher certificate \\
topic 4 & student class give rating lesson feedback time parent level bad \\
topic 5 & tutor account student lesson video group profile share bank paypal \\
\bottomrule
\end{tabular}}
\caption{\label{font-table} The first five topics of the starting model for the Reddit dataset. Both the new and old systems used the same starting model.}
\label{Table.main3}
\end{table}

\begin{table}[htb]
\centering
\resizebox{0.8\textwidth}{!}{
\begin{tabular}{p{2cm}p{12cm}}
\toprule
{\bf Topic} & {\bf Top 10 words} \\
\midrule
topic 1 & climate action change cop people glasgow world today day amp \\
topic 2 & party vote tory labour johnson boris government mps paterson sleaze \\
topic 3 & amp emissions countries climate carbon energy gas global coal fuel \\
topic 4 & charge pay people covid money case work claim give email \\
topic 5 & game great play team today enjoy amp fun win time \\
\bottomrule
\end{tabular}}
\caption{\label{font-table} The first five topics of the starting model for the Twitter dataset. Both the new and old systems used the same starting model.}
\label{Table.main4}
\end{table}

\begin{table}[htb]
\centering
\resizebox{0.95\textwidth}{!}{
\begin{tabular}{p{4cm}p{4cm}p{4cm}p{4cm}p{4cm}p{4cm}}
\toprule
{\bf Operation} & {\bf New system/Reddit} & {\bf New system/Twitter} & {\bf Old system/Reddit} & {\bf Old system/Twitter} & {\bf Total}\\
\midrule
{\bf Delete words} & 74 & 205 & 198 & 77 & 554\\
{\bf Reorder words} & 118 & 152 & 139 & 103 & 512\\
{\bf Add own words} & 14 & 45 & 32 & 24 & 115\\
{\bf Add suggested words} & 7 & 49 & / & / & 56\\
{\bf Delete document} & 12 & 37 & 20 & 12 & 81\\
{\bf Split topic} & 0 & 3 & 0 & 4 & 7\\
{\bf Merge topics} & 0 & 1 & 0 & 1 & 2\\
\bottomrule
\end{tabular}}
\caption{\label{font-table} The usage of refinements for each subject group. ``New system/Reddit'' indicates the use of new system with the Reddit dataset.}
\label{Table.main5}
\end{table}

\clearpage
\onecolumn
\section*{Appendix C: Tables for the use case one (section 4.3)}
\label{sec:appendix3}

\setcounter{table}{0}
\renewcommand{\thetable}{C\arabic{table}}

\begin{table}[htb]
\centering
\resizebox{0.8\textwidth}{!}{
\begin{tabular}{p{5cm}p{14cm}}
\toprule
{\bf Label} & {\bf Top 10 words} \\
\midrule
{\bf Reasons to join or leave a platform} & company, teacher, job, work, pay, people, make, money, online, think \\
\midrule 
{\bf Miscommunication with platform management} & know, anyone, video, tutor, help, ask, let, please, apply, interview \\
\bottomrule
\end{tabular}}
\caption{\label{font-table} Relevant topics identified by LDA, but missed by Model 16.}
\label{Table.main6}
\end{table}

\begin{table}[htb]
\centering
\resizebox{0.8\textwidth}{!}{
\begin{tabular}{p{5cm}p{14cm}}
\toprule
{\bf Label} & {\bf Top 10 words} \\
\midrule
{\bf Minimum and living wages} & pay, work, company, teacher, make, hour, rate, wage, job, time \\
\midrule 
{\bf Nationalities of students} & china, people, country, live, chinese, student, american, america, world, government \\
\midrule 
{\bf A restriction on what tutors can wear while tutoring} & wear, background, shirt, blue, eat, light, use, bao, love, dino \\
\midrule 
{\bf Chinese new law on out-of-school tutoring} & teacher, china, parent, chinese, company, foreign, school, class, new, student \\
\midrule 
{\bf Differences between a professional tutor and a community tutor} & student, lesson, teacher, tutor, use, english, learn, language, make, conversation \\
\midrule 
{\bf Discussions about Facebook groups created by platforms' management} & post, group, find, facebook, link, see, teacher, use, share, name \\
\midrule 
{\bf Tutors express anger and dissatisfaction with the platform} & people, make, good, fuck, post, shit, go, feel, na, sorry \\
\bottomrule
\end{tabular}}
\caption{\label{font-table} Relevant topics identified by Model 16, but missed by LDA.}
\label{Table.main7}
\end{table}

\begin{table}[htb]
\centering
\resizebox{0.8\textwidth}{!}{
\begin{tabular}{p{5cm}p{7cm}p{7cm}}
\toprule
{\bf Label} & {\bf Top 10 words (LDA)} & {\bf Top 10 words (Model 16)} \\
\midrule
{\bf Bookings and working hours} & week, hour, day, time, book, class, slot, schedule, open, month & class, week, time, day, hour, book, student, schedule, month, slot \\
\midrule 
{\bf Rating system} & student, teacher, work, class, give, rating, really, think, month, company & student, rating, call, minute, reservation, time, hour, ph, tutor, week \\
\midrule 
{\bf Teaching materials and methods} & student, use, lesson, question, word, ask, say, make, learn, conversation & word, use, game, read, student, slide, play, question, lesson, sentence \\
\midrule 
{\bf Technical issues} & class, minute, student, call, time, show, start, late, happen, reservation & use, issue, work, app, internet, problem, try, phone, computer, test \\
\midrule 
{\bf Payment} & lesson, pay, time, student, hour, teacher, rate, tutor, base, minute & class, hour, pay, per, bonus, teach, minute, base, lesson, student  \\
\midrule 
{\bf Experiences with kids in class} & kid, teach, level, well, old, year, think, really, feel, say & student, class, kid, teach, parent, time, lesson, give, teacher, make \\
\midrule 
{\bf Hiring process} & email, send, message, say, get, group, reply, back, try, see & interview, apply, video, demo, hire, good, application, new, company, process \\
\midrule 
{\bf Job requirements} & english, native, live, country, speaker, language, china, work, american, non & teach, online, english, native, tefl, company, work, experience, teacher, school \\
\midrule 
{\bf Bank transfers and transaction fees} & rating, account, pay, demo, use, bank, test, paypal, payment, internet & account, bank, pay, paypal, use, payment, transfer, payoneer, fee, money \\
\bottomrule
\end{tabular}}
\caption{\label{font-table} Relevant topics identified by both LDA and Model 16}
\label{Table.topics}
\end{table}

\clearpage
\onecolumn
\section*{Appendix D: Tables for the use case two (section 4.4)}
\label{sec:appendix4}

\setcounter{table}{0}
\renewcommand{\thetable}{D\arabic{table}}

\begin{table}[htb]
\centering
\resizebox{0.8\textwidth}{!}{
\begin{tabular}{p{5cm}p{14cm}}
\toprule 
 {\bf Category} & {\bf Definition}\\
\midrule
{\bf Market and industrial opportunities} & The potential of the invention to enter existing markets\\
\midrule
{\bf Cost and efficiency} & Reduction of production costs associated with more efficient processes\\
\midrule
{\bf Increasing compound yields} & Improvements in compound productivity based on novel processes\\
\midrule
{\bf Upscaling production} & Taking production to the commercial level\\
\midrule
{\bf Scientific advancements} & Contribution to knowledge production\\
\midrule
{\bf Environmental sustainability} & Contribution to environmental quality and preservation\\
\midrule
{\bf Human health} & Improvements in the quality of human health\\
\midrule
{\bf Food security} & Avoiding competition with human food sources\\
\midrule
{\bf Animal health} & Improvements in the quality of animal health\\
\bottomrule
\end{tabular}}
\caption{\label{font-table} categories table from \cite{ribeiro2020private}.}
\label{Table.main8}
\end{table}

\begin{table}[htb]
\centering
\resizebox{0.8\textwidth}{!}{
\begin{tabular}{p{5cm}p{14cm}}
\toprule 
 {\bf Category} & {\bf Final Concept Words}\\
\midrule
{\bf Market and industrial opportunities} & market, commercial, value, exists\\
\midrule
{\bf Cost and efficiency} & Reduction of production costs associated with more efficient processes\\
\midrule
{\bf Increasing compound yields} & compound, yield, increasing, plant\\
\midrule
{\bf Upscaling production} & production, improve, scale, large\\
\midrule
{\bf Scientific advancements} & advancement, benefit, filed\\
\midrule
{\bf Environmental sustainability} & renewable, sustainable, energy\\
\midrule
{\bf Human health} & health, human, patient, cancer, disease\\
\midrule
{\bf Food security} & food, security, supply, preparation, chain\\
\midrule
{\bf Animal health} & animal, health, disease\\
\bottomrule
\end{tabular}}
\caption{\label{font-table} Final concept words for each pre-defined category.}
\label{Table.main9}
\end{table}

\begin{table}[htb]
\centering
\resizebox{0.8\textwidth}{!}{
\begin{tabular}{p{5cm}p{14cm}}
\toprule 
 {\bf Category} & {\bf Topic (Top 10 words)}\\
\midrule
{\bf Market and industrial opportunities} & healthy, improved, serum, level, variant, commercial, value, concentration, woman, sample\\
\midrule
{\bf Cost and efficiency} & efficiency, increase, cost, increasing, efficient, enhance, reaction, amplification, low, target\\
\midrule
{\bf Increasing compound yields} & plant, tolerance, improved, compound, yield, marker, increasing, herbicide, soybean, resistance\\
\midrule
{\bf Upscaling production} & improve, improved, production, activity, antibody, enzyme, property, polypeptide, stability, expression\\
\midrule
{\bf Scientific advancements} & benefit, fold, effective, greater, composition, field, size, skin, provide, material\\
\midrule
{\bf Environmental sustainability} & renewable, energy, product, source, diesel, produced, carbon, lipid, sustainable, fuel\\
\midrule
{\bf Human health} & disease, patient, effective, cancer, amount, antibody, human, treatment, administering, subject\\
\midrule
{\bf Food security} & chain, healthcare, ge, mm, food, preparation, light, healthy, column, region\\
\midrule
{\bf Animal health} & health, animal, disease, effective, national, institute, dose, determined, administration, amount\\
\bottomrule
\end{tabular}}
\caption{\label{font-table} Focused topics identified by the initial model.}
\label{Table.main10}
\end{table}

\makeatletter
\setlength{\@fptop}{0pt}
\makeatother

\begin{table}[htb]
\centering
\resizebox{0.8\textwidth}{!}{
\begin{tabular}{p{5cm}p{14cm}}
\toprule 
 {\bf Category} & {\bf Topic (Top 10 words)}\\
\midrule
{\bf Market and industrial opportunities} & value, commercial, market, industrial, exists, business, global, adding, hepcidin, year\\
\midrule
{\bf Cost and efficiency} & efficiency, improve, increase, increasing, cost, enhance, reaction, efficient, amplification, low\\
\midrule
{\bf Increasing compound yields} & plant, efficiency, compound, yield, increasing, transformation, improved, gene, resistance, increased\\
\midrule
{\bf Upscaling production} & improve, production, improved, improvement, improving, higher, activity, greater, expression, enzyme\\
\midrule
{\bf Scientific advancements} & benefit, provide, field, improved, skin, cleaning, enzyme, surface, cellulase, material\\
\midrule
{\bf Environmental sustainability} & renewable, energy, reducing, biofuels, source, diesel, biofuel, carbon, lipid, product\\
\midrule
{\bf Human health} & disease, patient, cancer, diagnosis, effective, human, liver, lung, clinical, antibody\\
\midrule
{\bf Food security} & chain, sample, food, preparation, assay, light, supply, improved, donor, culture\\
\midrule
{\bf Animal health} & health, animal, medical, disease, care, healthy, nutrition, population, clinical, bethesda\\
\bottomrule
\end{tabular}}
\caption{\label{font-table} Focused topics identified by Model 23.}
\label{Table.main11}
\end{table}

\makeatletter
\setlength{\@fptop}{0pt}
\makeatother

\end{document}